\documentclass{llncs}

\usepackage{makeidx}  

\usepackage{graphics, graphicx, xcolor}
\usepackage{amsmath}
\usepackage{cite}

\usepackage{hyperref}
\hypersetup{
hidelinks, colorlinks, linkcolor={red},citecolor={blue}, urlcolor={blue}, backref=section
}

\usepackage{times}

\begin{document}

\textfloatsep=5pt
\intextsep=3pt

%
%
\title{Robust Landmark Detection for Alignment of Mouse Brain Section Images}
\titlerunning{Robust Landmark Detection for Mouse Brain Images}  
%
\author{Yuncong Chen\inst{1}, David Kleinfeld \inst{2}, Martyn Goulding\inst{3} \and Yoav Freund\inst{1}}
\authorrunning{Yuncong Chen et al.} 
%
%
\institute{UCSD CSE Department
\and UCSD Physics Department
\and Salk Institute for Biological Studies}

\maketitle              
\begin{abstract}

Brightfield and fluorescent imaging of whole brain sections are
fundamental tools of research in mouse brain study.  As
sectioning and imaging become more efficient, there is an increasing
need to automate the post-processing of sections for alignment and
three dimensional visualization. There is a further need to facilitate
the development of a digital atlas, i.e. a brain-wide map annotated
with cell type and tract tracing data, which would allow the automatic
registration of images stacks to a common coordinate
system. Currently, registration of slices requires manual
identification of landmarks. In this work we describe the first steps
in developing a semi-automated system to construct a histology atlas
of mouse brainstem that combines atlas-guided annotation,
landmark-based registration and atlas generation in an iterative
framework.
We describe an unsupervised approach for identifying and matching
region and boundary landmarks, based on modelling texture. Experiments
show that the detected landmarks correspond well with brain
structures, and matching is robust under distortion. These results
will serve as the basis for registration and atlas building.


\end{abstract}

\section{Introduction}

In this paper we describe a method for the automatic detection of
landmarks in histology images of mouse brain sections. The purpose is
to facilitate image registration and atlas generation. By aligning
images of nissl and fluorescent-stained brain sections, we aim to
create a digital atlas for the mouse brainstem, which incorporates
cell type, tract tracing and other physiological data. Unlike most
brain atlases which are intended to be viewed by a human, the atlas we
aim to build is intended to be used by a computer. Specifically, the
goal is to represent the landmarks in a datastructure that would be
used by an algorithm to automatically align a whole stack to a common
coordinate system.  The novelty of the work described here is the use
of unsupervised learning to find regions with distinct texture and
clear boundaries. This will reduce the manual work needed to identify
reliable landmarks.

Figure \ref{fig:TextonHistComparison} shows a typical image of nissl-stained mouse brainstem section. Although the brainstem does not have salient edges like the cortex, it contains many compact neuron clusters (nuclei) and striated regions (fiber tracts). Both types of structure have distinct texture that can be detected and modelled.

We use semi-supervised learning to create a library of landmark detectors based on texture and shape. They can be applied to identify landmarks from new images and also be updated by incorporating new data. This is at the center of our vision of integrating atlas-guided annotation, landmark-based registration and atlas generation into an iterative framework.

The remaining of this paper is organized as the follows. Section 3 describes how we represent textures using Gabor filters and superpixels. Section 4 and 5 explains how we identify potential landmark regions based on texture distinctiveness. Section 6 describes how we detect stable boundary segments to complement region landmarks. Section 7 describes how landmarks are matched between sections.



\section{Related Work}

%


Existing work on automatically reconstructing 3D volume from histology section include the Allen Mouse Brain Atlas \cite{lein2007genome, yushkevich20063d} and the Waxholm Space \cite{johnson2010waxholm}, both of which use intensity-based methods such block-matching \cite{ourselin2001reconstructing, roberts2012toward} and mutual information maximization. Landmark-based methods take advantage of details in the histology image using descriptors such as SIFT\cite{sun2012nearly} and binarized gradient orientation histogram \cite{kurkure2011landmark}.


\section{Representing Texture using Histograms of Gabor Textons} 

%
%
%

Images are first filtered using Gabor filters\cite{jain1990unsupervised} with 9 orientations and 11 scales, resulting in 99 dimensional feature vectors. We reduce the data by quantizing these feature vectors into 100 clusters using K-Means. Clusters with close centroids are then merged, and the final cluster centroids are the textons. In our experiment, we have 14 textons.

\begin{figure}
\centering
	\includegraphics[width=.8\textwidth]{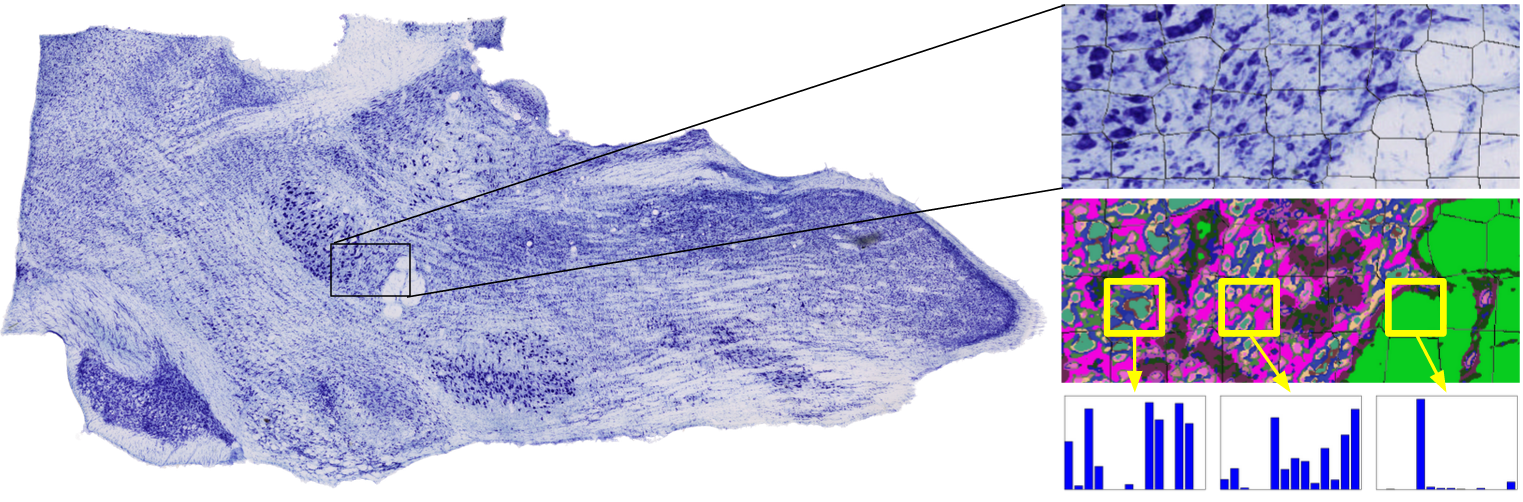}
	\caption{Left: A typical image of mouse brainstem section. Right top: part of the original image with superpixels overlaid. Middle: texton map of the same area. Bottom: texton histograms of three superpixels with different textures.}
	\label{fig:TextonHistComparison}
\end{figure}

We further reduce the data by describing texture at the level of superpixels. For each superpixel, texture is represented by the histogram of the textons it contains (Figure \ref{fig:TextonHistComparison}). We denote the texton histogram of the $i$'th superpixel by $h_i$. Superpixels are obtained using the SLIC algorithm \cite{achanta2012slic}.

Since different sections may be oriented at different angles, 
we need the features to be rotation-invariant. For this purpose, before doing K-Means, we compute the energy distribution over orientations for each Gabor feature vector, and shift all feature vectors such that the modes of the directional energy distributions are aligned. In this way, the texture is decoupled from directionality.

\section{Evaluating Region Significance using Statistical Test}

We define a region landmark as a connected set of superpixels. Before describing how to find significant regions, we need a scoring function that evaluates the significance of a region. We propose the significance score, $F(S)$, of a region $S$, to be a linear combination of three terms. 

The first term is the distinctiveness of this region's texture, or in other words, how different its texture is from the surrounds. Since textures are represented as histograms, we formulate this problem as a statistical test on whether the region's average texton histogram and the texton histograms of the surrounding superpixels are sampled from the same distribution. We employ the chi-squared test for independence and the resulting p-value serves as a quantitative measure of the distinctiveness. The smaller the p-value, the more significant the region is. Since a region often has neighbours with different textures, the test is made between the region's histogram and that of every surrounding superpixel. To be conservative, the largest p-value among all tests is used. Denote the set of surrounding superpixels by $T(S)$, the first term is,
$$F^{cont}(S) = - \max_{j\in T(S)} pval(h_S, h_j)$$
where $pval(\cdot,\cdot)$ is the p-value of applying the chi-square test for independence to two histograms.

The second term is the texture homogeneity within the region, defined as the mean p-value of chi-square tests between all superpixels in the region and the region's average:
$$F^{coh}(S) = \frac{1}{|S|} \sum_{i\in S} pval(h_S, h_i)$$

The third term stems from the fact that most region landmarks have compact shapes. A common measure for the compactness of a closed curve is the isoperimetric quotient, defined as the enclosing area divided by the square of the circumference. We use the number of surrounding superpixels as a proxy for the circumference, and the number of superpixels in the region as that for the area. 
$$F^{comp}(S) = \frac{|T(S)|}{|S|^2}$$

The overall significance score $F(S) = w_1 \cdot F^{cont}(S) + w_2 \cdot F^{coh}(S) + w_3 \cdot F^{comp}(S)$, with weights chosen empirically.

\section{Finding Significant Regions with Region Growing and Clustering}

\begin{figure}
\centering
	\includegraphics[width=.8\textwidth]{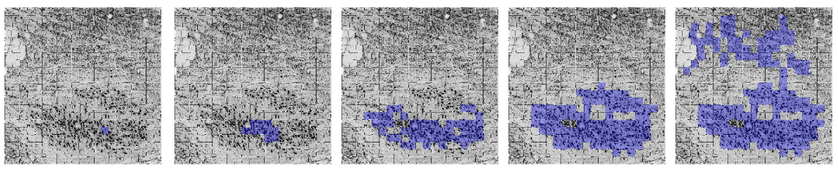}
	\caption{Regions during growing, from left to right, at iteration 1, 10, 50, 96 (most significant), 150 (last iteration).} 
	\label{fig:RegionGrowing}
\end{figure}

To find potential regions, we perform region growing for every superpixel. Starting with a singleton containing the seed superpixel, this greedy procedure considers all neighbours of the current set, and iteratively adds the one whose texture is closest to the current region's texture, measured by the $\chi^2$ distance between texton histograms. The growing stops when the region reaches 10\% of the total area. Eventually, the region at the point when the significance score is the largest is returned. Figure \ref{fig:RegionGrowing} shows examples of this procedure.

\begin{figure}
\centering
	\includegraphics[width=.8\textwidth]{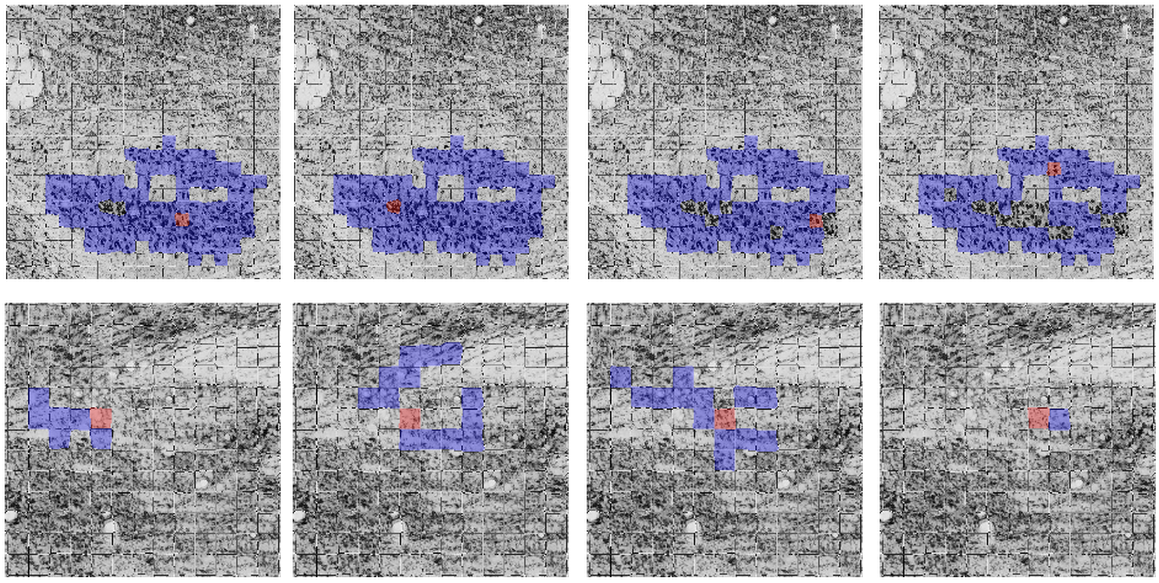}
	\caption{Top row: the region proposals of four different seeds (in red) in the facial motor nucleus. They are very similar. Bottom row: the proposals of four neighboring seeds in an inhomogeneous region. They are very random.} 
	\label{fig:ConsistentProposals}
\end{figure}

We call the set of superpixels that grows from a seed superpixel $k$ the \textit{region proposal} of the seed, denoted by $S_k$. If a region has distinct texture, the proposals of superpixels inside this region should be very similar, while proposals from within an inhomogeneous region are random (see Figure \ref{fig:ConsistentProposals}). In other words, the region proposals form clusters. The denser a cluster is, the more distinctive the region it represents. Using Jaccard index as pairwise distance, we group the region proposals using hierarchical clustering. Then within each group whose size is large enough, we select the proposal with the highest significance score to be the representative of that group. All representative proposals are ranked according to their significance scores. 


\section{Identify Robust Boundaries by Region Consensus}

\begin{figure}
	\centering
	\includegraphics[width=.8\textwidth]{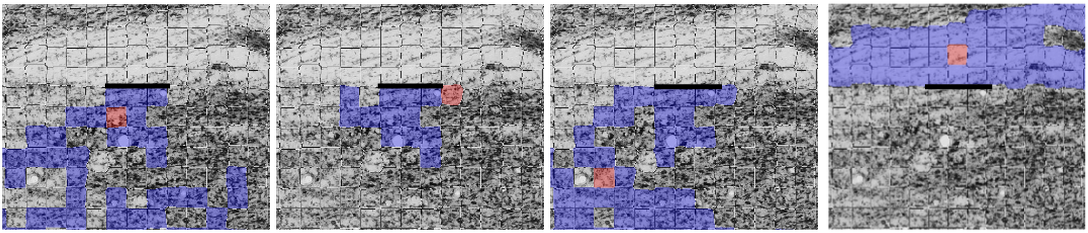}
	\caption{The three region proposals in an inhomogeneous area are not consistent as a whole, but they all agree on a robust boundary segment (highlighted).}
	\label{fig:RobustBoundaryExample}
\end{figure}

Sometimes a region gradually transitions into neighboring texture on one side, but has a clear boundary on the other side. The method described in the previous section may not capture this inhomogeneous region, but the open boundary is nonetheless a perfect landmark. Figure \ref{fig:RobustBoundaryExample} shows such an example. Notice how much the proposals from seeds in such region vary, but many of them still agree on the clear boundary. This motivates a consensus-based approach for evaluating boundary robustness, in which each region proposal votes for the segments on its boundary.

We represent a boundary segment between two superpixels by an ordered tuple $(i,j)$, where $i$ is the interior superpixel and $j$ is the exterior superpixel. We also denote by $\delta S_k$, the set of segments on the boundary of a region proposal $S_k$. 
The vote a region proposal casts to a boundary segment depends on how contrasty the segment is, measured by the texture histogram distance between the region's average and the segment's exterior superpixel. We call the set of superpixels that vote for a segment the \textit{supporter set} of the segment, and denote it by $R_{(i,j)} = \{k: (i,j) \in \delta S_k\}$. The total score received by a segment $(i,j)$ is then: 
$$ b_{(i,j)} = \sum_{k \in R(i,j)} \chi^2(h_{S_k}, h_j)$$

\begin{figure}
	\includegraphics[width=\textwidth]{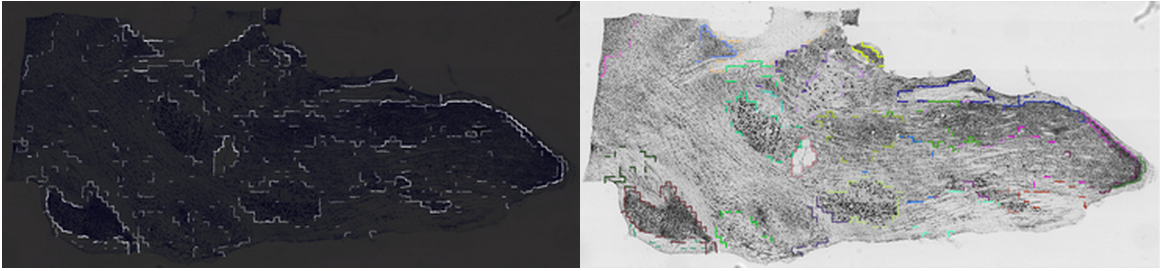}
	\caption{Left: thresholded boundary vote map. Right: grouped boundary segments}
	\label{fig:BoundaryMap}
\end{figure}

We discard segments whose vote is lower than a threshold. Figure \ref{fig:BoundaryMap} shows a vote map. Instead of modelling individual segments, we combine segments with similar supporter sets into groups, again using hierarchical clustering with Jaccard index between supporter sets as pairwise similarity. Each segment group represents a boundary. All boundaries are ranked according to the total vote received by its segments. 

\section{Matching Landmarks from Different Sections}

By representing region landmarks using closed boundaries and merging coinciding boundaries, we unify the two types of landmarks into a single set consisting of both open and closed boundaries. In order to use the landmarks for registration, correspondences must be made between them. In this section, we design a distance function for comparing two boundaries. This function is a weighted combination of the differences in four aspects, with weights chosen empirically:
\begin{align*}
D(\mathbf{B}_1, \mathbf{B}_2) = & ~w^{int} D^{int}(\mathbf{B}_1, \mathbf{B}_2) + w^{shape} D^{shape}(\mathbf{B}_1, \mathbf{B}_2)  \\
+ & ~w^{ext} D^{ext}(\mathbf{B}_1, \mathbf{B}_2) 
+ w^{loc} D^{loc}(\mathbf{B}_1, \mathbf{B}_2) 
\end{align*}

The first term is the difference of interior textures. For a region landmark $S$, the interior texture is simply the average texton histogram $h_S$. For a boundary $B$, the interior texture is the average texton histogram of the union of all segments' supporter sets. Denote this union by $Q_B = \cup_{(i,j)\in B} R_{(i,j)}$, then $h^{int}_B = h_{Q_B}$. The difference is computed using the $\chi^2$ distance:
$$D^{int}(\mathbf{B}_1, \mathbf{B}_2) = \chi^2(h^{int}_{B_1}, h^{int}_{B_2})$$

The second term measures the similarity of boundary shapes. We reduce the boundary to a point set consisting of midpoints of the segments. The shape distance between two point sets can be computed using shape context descriptors\cite{belongie2000shape}. The shape context descriptors characterize the organization of other points around each point using a histogram. Two sets of points are matched by finding the minimum bipartite matching, where the edge weights are the chi-square distances between shape context descriptors. The Hungarian algorithm is used to find the minimum matching. The average cost of this matching, is used as the second term:

$$D^{shape}(\mathbf{B}_1, \mathbf{B}_2) = \frac{1}{|M|}\sum_{(i,j), (p,q) \in M(B_1,B_2)} \chi^2(c_{i,j}, c_{p,q})$$ where $M(B_1,B_2)$ is the minimum matching, $c_{i,j}$ and $c_{p,q}$ are the shape context descriptors of segment $(i,j)$ and segment $(p,q)$ respectively.

After the matching is made, we compute the third term, defined as the total distance between exterior textures of all matched segments:
$$D^{ext}(\mathbf{B}_1, \mathbf{B}_2) = \sum_{(i,j), (p,q) \in M(B_1,B_2)} \chi^2(h_j, h_q)$$

The fourth term measures the spatial proximity. This is the thresholded Euclidean distance between the center of mass of the boundaries' midpoint sets:
$$D^{loc}(\mathbf{B}_1, \mathbf{B}_2) = \max(0, \| m_{B_1}, m_{B_2} \|_2 - l)$$
where $m_{B_1}$ and $m_{B_2}$ are the center of mass of boundaries $B_1$ and $B_2$, and $l$ is a tolerance within which the position deviation is not penalized (set to 1 mm in our experiment).



Using this distance function, we compute the pairwise distances for two sets of landmarks detected from different sections. Two landmarks are matched if they are simultaneously the closest landmark to each another.

%
%

\section{Experiments}

\subsection{Comparison to Human Labelling}

\begin{figure}
	\includegraphics[width=\textwidth]{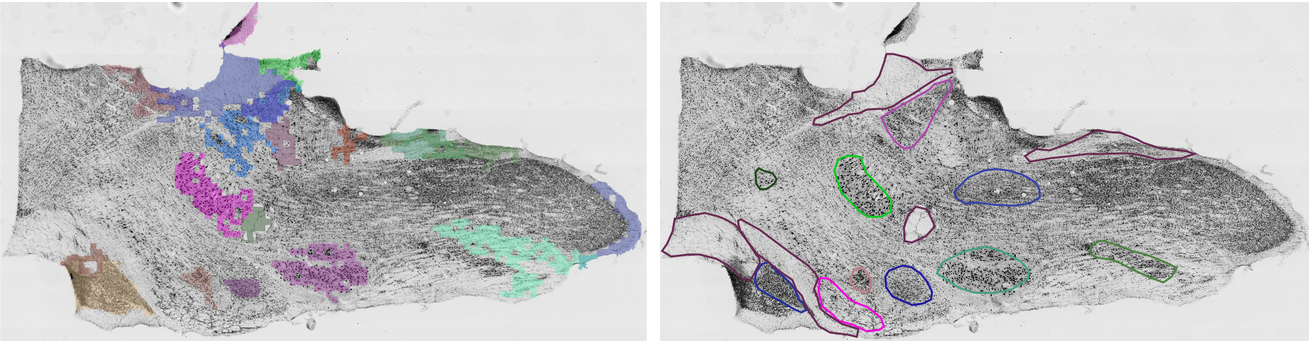}
	\caption{Left: top 20 landmarks detected by the algorithm. Right: human labelling for recognizable nuclei and white matter.}
	\label{fig:TopRegions}
\end{figure}

We test our algorithm on a series of 30 section images of a nissl-stained mouse brainstem. The images are scanned at 2 microns per pixel, showing individual neuronal cell bodies. Both types of landmarks are detected on all images. For each image we use the top 20 closed boundary and the top 10 open boundary. Figure \ref{fig:TopRegions} shows the landmarks detected from one image. Also shown is a version created by a human labeller who annotated for nuclei and fiber tracts with the help of a printed atlas. Most significant structures from the human labelling are detected by the algorithm.


\subsection{Robustness of Matching}

\begin{figure}
	\includegraphics[width=\textwidth]{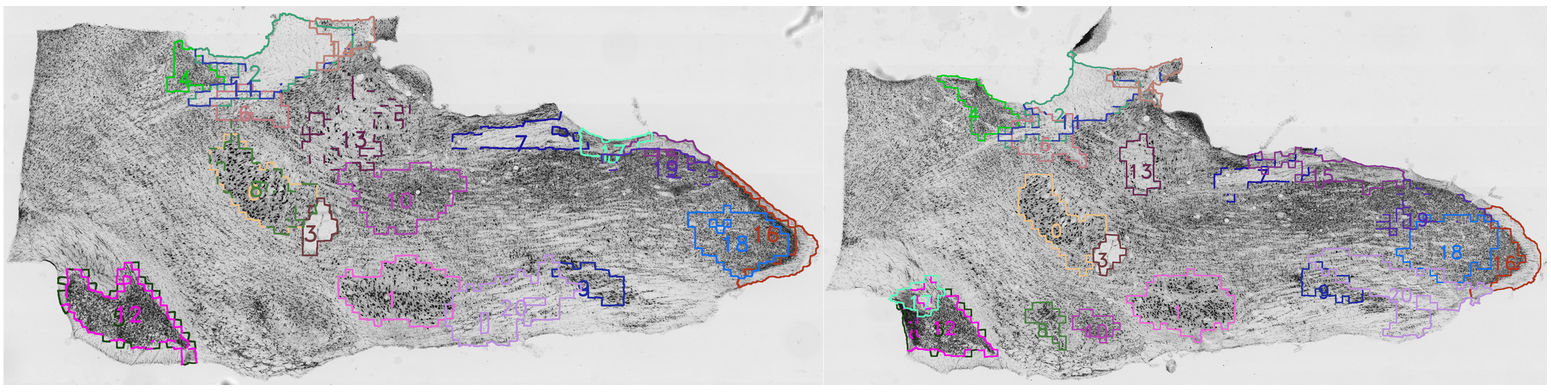}
	\caption{Landmark matching example. Matched landmarks are marked with the same color and number. Note that matchings are found based only on texture and shape. Matching 13 is made possible by modelling both open boundaries and close boundaries. Landmarks such as 12 and 9 show considerable shape change, but are still matched.}
	\label{fig:LandmarkMatch}
\end{figure}

The landmark matching algorithm is applied to all 30 pairs of consecutive images. In order to test the robustness of matching under large displacement, we remove the spatial proximity term from the landmark distance function, leaving only the texture and shape terms. A human evaluator then judges whether a matching is correct, partially correct, or incorrect. Among all 166 matchings returned by the algorithm, 106 are correct (63\%), 34 are partially correct (22\%) and 26 are wrong (15\%). One example is shown in Figure \ref{fig:LandmarkMatch}. This demonstrates the effectiveness of texture modelling. Even though some landmarks change shape significantly, the algorithm still finds the correct matchings.

%
%
%
%
%
%
%
%
%
%
%
%
%
%
%

%

\section{Conclusion and Future Work}

In this paper we described algorithms for detecting region and boundary landmarks from histology images of mouse brainstem. Region proposals are grown from superpixels, based on which significant regions such as nuclei and fiber tracts are identified using clustering. Detected regions are shown to correspond well with real anatomical structures. To complement region landmarks, robust boundary segments are also found by consensus voting. Landmarks from different sections can be matched using a distance function that is robust under distortion.

Our next step is to use the matched landmarks for intra-specimen registration, and further co-register multiple specimens to generate the atlas. Meanwhile, we also plan to include human feedback in the learning loop. Although the advantage of our approach is reducing human supervision, it is important for human experts to correct wrong labellings made by the algorithm in order to learn accurate models for landmarks. Instead of requiring them to label every region of interest, our method only needs them to validate our library of detectors. Once those have been validated, the process is completely automatic.

%


%
%

\bibliography{bibfile}
\bibliographystyle{splncs03}

%
%

\end{document}